# Determining Vertical Displacement of Agricultural Areas Using UAV-Photogrammetry and a Heteroscedastic Deep Learning Model

**Wojciech Gruszczyński *, Edyta Puniach, Paweł Ćwiąkała and Wojciech Matwij**

AGH University of Krakow, Faculty of Geo-Data Science, Geodesy, and Environmental Engineering, Mickiewicza 30, 30-059 Krakow, Poland; epuniach@agh.edu.pl (E.P.); pawelcwi@agh.edu.pl (P.Ć.); matwij@agh.edu.pl (W.M.)
* Correspondence: wgrusz@agh.edu.pl

## Highlights

**What are the main findings?**

- The U-Net model predicts elevation corrections and quantifies their uncertainty, enabling subsidence determination with minimal influence of vegetation.
- Among the tested conventional ground filters, Adaptive TIN achieved the best performance subsidence determination task.

**What is the implication of the main finding?**

- The proposed approach provides slightly better performance than conventional ground filters in determining subsidence over agricultural areas.
- The accuracy and data density achieved with the U-Net model are sufficient for reliable assessment of subsidence-related risks in agricultural areas.

## Abstract

This article introduces an algorithm that uses a U-Net architecture to determine vertical ground surface displacements from unmanned aerial vehicle (UAV)-photogrammetry point clouds, offering an alternative to traditional ground filtering methods. Unlike conventional ground filters that rely on point cloud classification, the proposed approach employs heteroscedastic regression. The U-Net model predicts the conditional expected values of the elevation corrections, aiming to reduce the impact of vegetation on determined ground surface elevations. Concurrently, it estimates the logarithm of the elevation correction variance, allowing for direct quantification of the uncertainty associated with each elevation correction value. The algorithm was evaluated using three metrics: the root mean square error (RMSE) of vertical displacements, the percentage of nodes with determined displacement values, and the percentage of outliers among those values. Performance was assessed using the technique for order of preference by similarity to ideal solution (TOPSIS) method and compared against several ground-filter-based algorithms across four datasets, each including at least two time intervals. In most cases, the U-Net-based approach demonstrated a slight performance advantage over traditional ground filtering techniques. For example, for the U-Net-based algorithm, for one of the test datasets, the RMSE of the determined subsidences was 6.1 cm, the percentage of nodes with determined subsidences was 80.5%, and the percentage of outliers was 0.2%. For the same case, the algorithm based on the next best model (SMRF) allowed an RMSE of 7.7 cm to be obtained; for 77.3% of nodes, the subsidences were determined; and the percentage of outliers was 0.3%.

## 1. Introduction

Ground surface displacements caused by underground mining operations lead to deformation. Traditionally, geodetic surveys aimed at monitoring the impact of underground mining are conducted with high precision along selected profiles, sometimes supplemented by clusters of points located around structures of particular significance. These profiles should be strategically selected to capture the maximum possible displacements and deformations. However, in practice, their placement is often dictated by the layout of roads. This approach results in two significant issues: a relatively small number of observed points and their often suboptimal placement from the perspective of surface monitoring. As a result, despite the high accuracy of such measurements, the surface displacement model and the associated risk assessment may significantly deviate from reality.

The advancements in remote sensing technologies for ground surface observation provide several significant alternatives to traditionally measured profiles. These include Aerial Laser Scanning (ALS), Terrestrial Laser Scanning (TLS), Interferometric Synthetic Aperture Radar (InSAR), and aerial photogrammetry, particularly unmanned aerial vehicle (UAV) photogrammetry. Each of these methods offers numerous advantages but also presents specific challenges and limitations. A detailed discussion of the characteristics of ALS, TLS, and InSAR goes beyond the scope of this article, and thus, the focus is placed on UAV-photogrammetry.

The advantages of UAV-photogrammetry for measuring ground surface displacements include high operational flexibility, relatively low equipment and survey costs for small areas, and high observation resolution, i.e., a small ground sampling distance (GSD). However, its primary drawback is that the accuracy of the determined displacement is highly dependent on surface coverage, particularly vegetation conditions.

Under favorable conditions, the accuracy of elevation determination for point clouds and associated vertical displacements is approximately 2–3 GSD, which is generally sufficient for practical applications. However, vegetation can significantly amplify errors. This can manifest as increased random errors (uncertainty in point cloud elevation determination) and systematic errors (where the point cloud reflects the elevation of vegetation rather than the actual ground surface).

Since vertical displacement determination relies on point cloud elevations from two UAV measurement series, the total vertical displacement error depends on the accuracy of ground elevation determination in both flights. The state of vegetation changes between surveys conducted over time due to natural growth, mowing, or crop rotation. Additionally, agricultural activities such as furrow placement and other small-scale surface modifications may alter the ground's microrelief.

These factors mean that simply subtracting the elevations of point clouds from two recorded states (measurement series) will often produce a significantly distorted representation of ground surface displacement. However, this does not render the data entirely unusable; rather, it limits and complicates its interpretation [1,2]. Reliable analysis is only feasible in areas where the impact of vegetation on point cloud elevations is comparable in both datasets (minimizing systematic errors) and minimal (reducing random errors). For this reason, most studies use ground-filtering-based classification to reduce the impact of vegetation before digital elevation model (DEM) construction [3–7].

A study by Lian et al. [3] also compared five widely used ground filtering algorithms in their ability to classify ground and non-ground points in mountainous areas above an ongoing underground mining operation. The comparison was made in relation to reference point cloud classification, and the metrics used were type I and II errors, as well as the total error. The best classification results in this instance were achieved by an algorithm based on Adaptive TIN (ATIN) [8]. The final accuracy of vertical displacements was assessed based on comparison to reference values determined using leveling of marked points. The achieved root mean square error (RMSE) of vertical displacement was ~16 cm.

Most existing algorithms designed to minimize vegetation impact on point cloud elevations have been developed for filtering data acquired using ALS [8–17]. These algorithms typically classify point cloud data into two categories: ground surface and other features (vegetation, buildings, etc.). Fundamentally, these algorithms distinguish ground surface points from others based on a threshold value for terrain slope or curvature. The underlying assumption is that lower values of these parameters correspond to the ground surface, while higher values indicate non-ground features.

In recent years, numerous studies have examined semantic segmentation of point clouds using machine-learning methods [18–23]. Although typically designed for multi-class segmentation, these models can be trained for ground-filtering tasks. Qin et al. [24] compared several such models [20–23] against classical ground-filtering algorithms [8,11,12,17]. The comparison highlighted several advantages of machine-learning-based approaches, including flexibility in mixed-terrain scenes, robustness in complex landforms, reduced sensitivity to dense outliers, and consistent computational efficiency. It also identified challenges, notably occasional micro-topographic errors and degraded cross-dataset generalization attributable to overfitting. Overall, their study concludes that applying machine-learning point-cloud segmentation algorithms to ground filtering is promising while emphasizing the need to address these limitations and outlining directions for future research.

While all of the aforementioned algorithms rely on classification, Gruszczyński et al. [25] proposed an alternative approach aimed at reducing the impact of vegetation. In this method, a U-Net neural network [26] (homoscedastic), a type of convolutional network [27], was used to correct point cloud elevations to better reflect the actual ground surface. A second U-Net, trained to estimate the uncertainty of these corrections, was then applied to select only points exceeding a predefined accuracy threshold. Together, these two networks implemented heteroscedastic regression of the height correction values. While effective in determining bare-earth elevations, this approach faced several challenges. The availability of training data was limited, and the results of the variance-estimating network were highly dependent on the performance of the network computing the elevation corrections. Training two interconnected networks required careful attention to generalization, as any overfitting in the first network negatively affected the second network's uncertainty predictions. Additionally, using two models potentially increased computational costs due to dual predictions, and the variance estimation network occasionally produced negative variance values, which are mathematically inconsistent.

In this study, these limitations were addressed by employing a single U-Net model that simultaneously predicts elevation corrections for a uniformed point cloud and, after specific transformations, estimates the uncertainty of these corrections. Compared to the previously described approach of Gruszczyński et al. [25], modifications were introduced in the data preprocessing method and the loss function. Additionally, the model was trained on a significantly larger dataset. Collectively, these improvements led to substantial advancements over the previous solution.

Moreover, the proposed approach extends previous work by enabling the determination of vertical displacements rather than just corrected terrain elevations. Its performance is evaluated against three widely used ground filtering algorithms—ATIN [8], Simple Morphological Filter (SMRF) [15], and Cloth Simulation Filter (CSF) [17]—to assess their relative effectiveness in mitigating the influence of vegetation on the estimation of vertical ground displacements in agricultural areas.

Section 2 describes the point cloud processing methods used to determine vertical displacement values, the datasets employed in this study, and the methodology for comparing results across different algorithms. Section 3 presents the results of these comparisons, while Section 4 provides a discussion of the findings. Finally, Section 5 synthesizes the conclusions drawn from the research.

## 2. Materials and Methods

Section 2.1 describes the proposed point cloud processing algorithm, enabling the application of U-Net to determine vertical displacements in agricultural areas. Section 2.2 outlines the parameter ranges used for ground filtering algorithms, as well as the data processing applied to determine vertical displacements. Section 2.3 provides an overview of the datasets used for training the U-Net and evaluating its performance in comparison to other algorithms.

### 2.1. U-Net-Based Algorithm

This subsection describes the proposed algorithm for utilizing the U-Net neural network to determine vertical ground surface displacements. The concept (Figure 1) is based on correcting the elevations of point cloud data to match the actual ground surface elevations while also predicting the uncertainty associated with these corrections. This approach allows only those segments of the corrected point clouds that correspond to ground surface elevations with relatively low uncertainty to be retained. The difference between these corrected elevations is then interpreted as the vertical displacement of the ground surface.

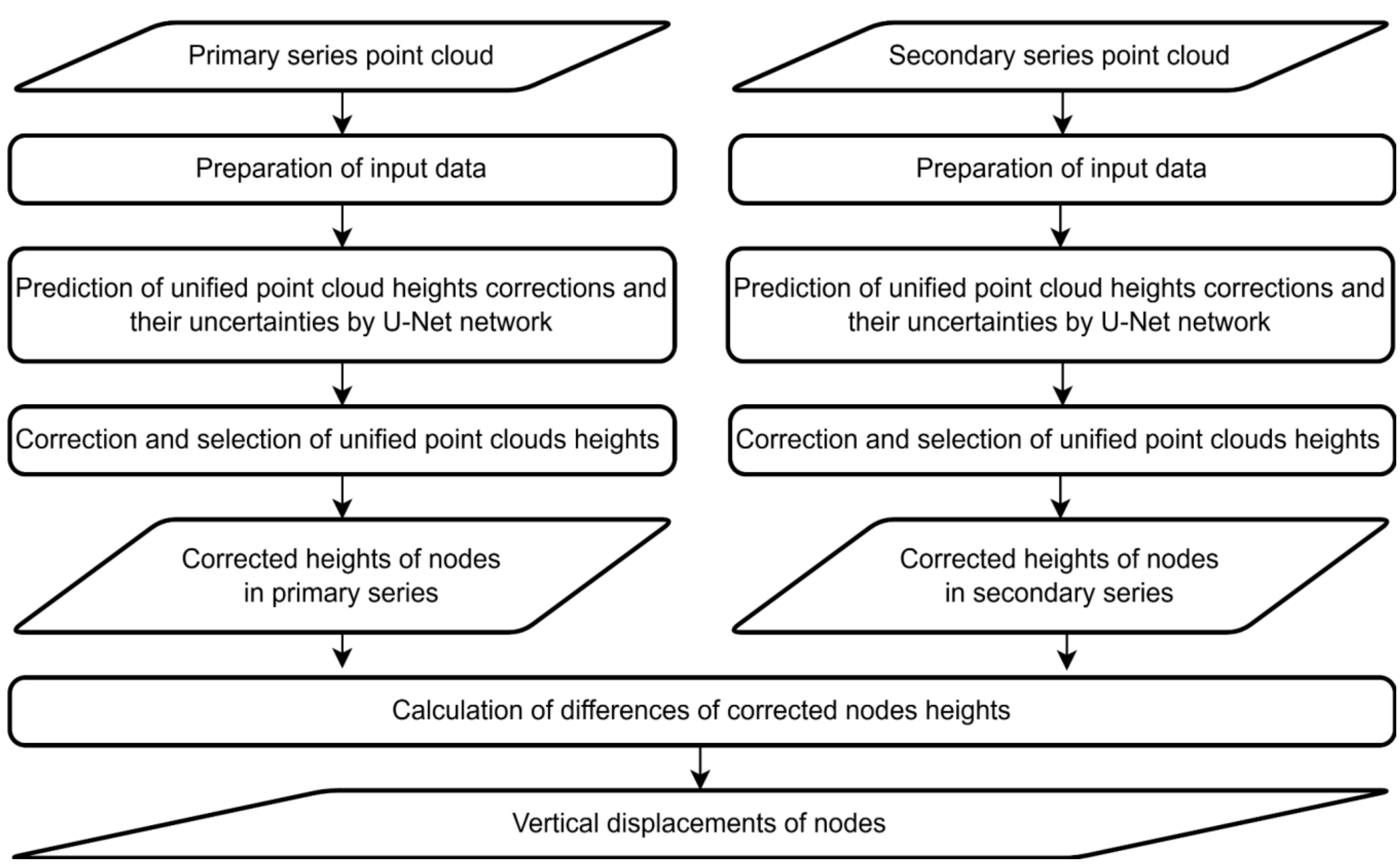

**Figure 1.** General data flowchart.

The subsections provide a detailed description of the processing of the point clouds into the input data for the U-Net and the method for determining reference correction values, as well as the structure and training process of the neural network. The section 2.1.4. explains in detail the procedure for determining vertical displacements based on the results generated by the neural network.

#### 2.1.1. Input Data Formulation

The process of formulating input data for the neural network based on the point cloud is illustrated in a simplified manner in the flowchart shown in Figure 2. In the first step, the area covered by the point cloud is divided into 5 cm × 5 cm grid cells. Each cell is assigned the elevation of the lowest point within its boundaries. The uniform density point cloud obtained through this process is then stored in the variable *uni*.

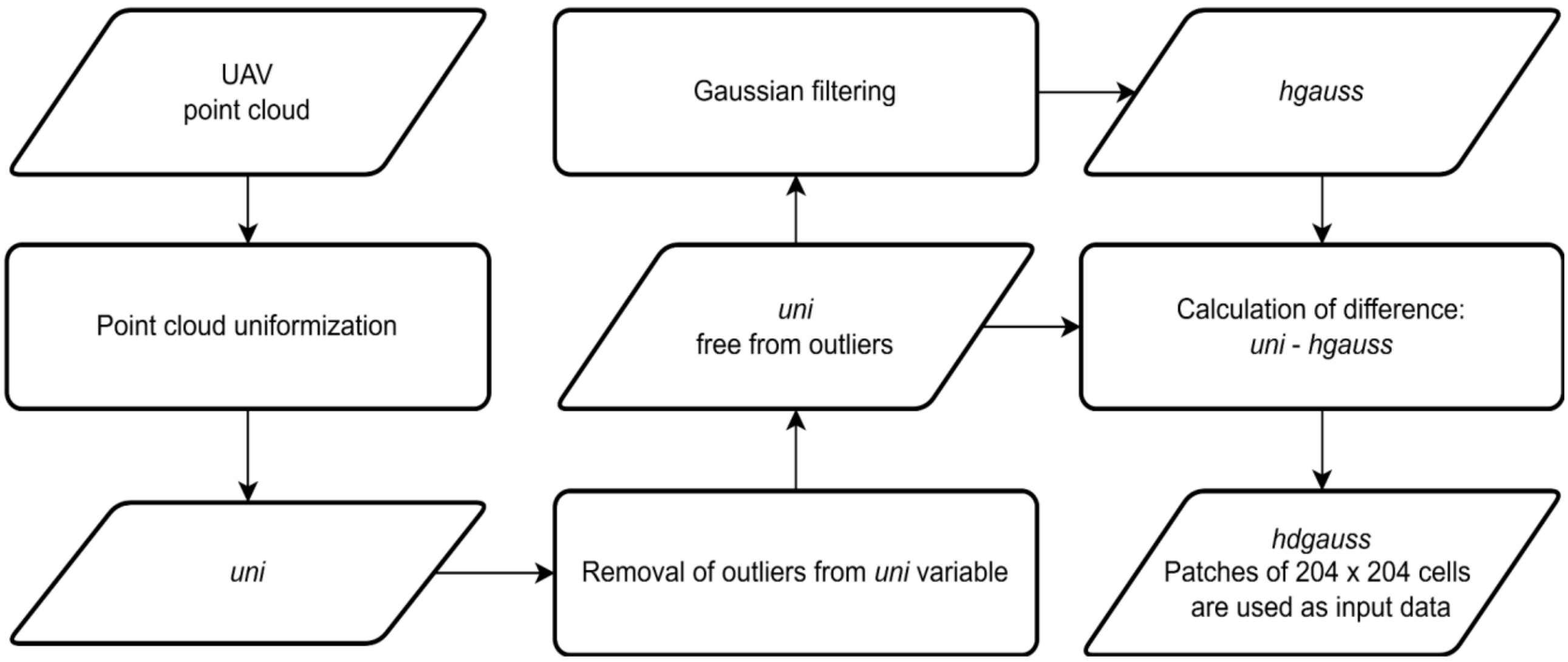

**Figure 2.** Formulation of input data flowchart.

Next, outlier observations—which may appear in the point cloud either significantly above or below the remaining points—are removed (Figure 3). This process is inspired by the outlier removal algorithm implemented in the SMRF algorithm. The parameter values of the outlier removal algorithm were selected through trial and error.

**Input:**

*uni* – uniformed point cloud

*slopethresh* – outlier detection threshold (e.g., 10 m)

*windowradius* – radius of morphological opening (e.g., 2 m)

**Process:**

1. Construct sparse grid (*hsparse*):
   For each grid cell (1 m × 1 m):
   *hsparse*(cell) = minimum elevation of *uni* points within cell
2. Fill gaps in *hsparse*:
   *hsparse* = inpaint(*hsparse, missing*)
3. Detect outliers above terrain:
   *dem* = morphological_open(*hsparse*, *windowradius*)
   *outliers* = |*dem* – *hsparse*| > *slopethresh*
4. Detect outliers below terrain:
   *dem* = morphological_open(–*hsparse*, *windowradius*)
   *outliers* = *outliers* OR (|*dem* – (–hsparse)| > *slopethresh*)
5. Replace outliers in *hsparse*:
   *hsparse* = inpaint(*hsparse,outliers*)
6. Compare *uni* against *hsparse*:
   For each grid cell:
   IF |*hsparse* – *uni*| > *slopethresh* THEN *uni*(cell) = missing
7. Interpolate missing values in *uni*:
   *uni* = inpaint(*uni,missing*)

**Output:**

*uni* free of outliers, suitable for further processing

**Figure 3.** Pseudocode of outlier removal algorithm. The function inpaint is used to interpolate missing or outlier values in the first variable, whereas morphological_open represents a morphological opening (erosion followed by dilation).

Next, the *uni* variable undergoes gaussian blurring with a filter size of 205 × 205 cm (41 × 41 cells) and a standard deviation of 50 cm (10 cells). The result of this operation is stored in the variable *hgauss*. The difference between *uni* and *hgauss* is then computed and stored in the variable *hdgauss*. Patches extracted from *hdgauss* are used as inputs for the neural network during both training and prediction.

2.1.2. Target Correction Values

The elevation data required to calculate target height correction values for training the neural network is derived from point clouds obtained through ALS and stored in Polish National Geodetic and Cartographic resources. This data is saved in the LAS format in accordance with the specification version 1.2 published in 2008 by the American Society for Photogrammetry and Remote Sensing. According to the information provided by the institution providing the ALS data, the correctness of point classification in the ALS point cloud is at least 95%.

Only points representing the last returns classified as the ground and buildings (i.e., points whose elevations should remain unchanged over time) were retained. The points classified as buildings were included as a precaution to ensure the algorithm operated correctly even in urbanized areas when determining subsidence. Before using the ALS point clouds for training, they were cleaned of potential classification and systematic errors. The outliers caused by measurement inaccuracies or terrain elevation changes due to land use modifications between ALS surveys and UAV flights were also removed.

To eliminate systematic elevation discrepancies, ALS point clouds were compared with elevation measurements obtained from global navigation satellite system (GNSS) real-time kinematic (RTK) surveys conducted on a regular grid. For each ALS flight strip, the median discrepancy between these elevations (ALS–GNSS RTK) was calculated, and the ALS point cloud was shifted by this value accordingly.

To eliminate outliers in the ALS point cloud, ALS elevation data after the shift was compared with measurements from GNSS RTK surveys. Additionally, ALS elevations were validated by comparing them with a digital surface model (DSM) derived from UAV-photogrammetry conducted during periods of negligible vegetation influence (late winter, provided there was no snow or early spring). If the discrepancy between ALS elevations and those from UAV-photogrammetry or GNSS RTK exceeded 20 cm, the corresponding section of the ALS point cloud was excluded.

The elevations from the cleaned ALS point cloud were interpolated onto 5 cm × 5 cm grid cells, aligned with cells used for generating the input values for the neural network. For each dataset used in network training, the differences between the elevations in the uniformed UAV-photogrammetry point cloud and the ALS elevations were calculated. These values represent the elevation of uni points above the ground surface and are referred to as corrections (to elevation) in the following sections. If the absolute value of these corrections exceeded 5 m, the corresponding grid cells (output image pixels) were excluded from training. These corrections serve as the targets during the training of the U-Net network.

#### 2.1.3. Neural Network and Its Training

To model the values of elevation corrections, a U-Net network with an input size of 204 × 204 and one channel was used. The depth of the encoder and decoder was set to 4. Because there was no zero padding applied at the inputs to the convolution layers, the output size had to be set to 20 × 20 and 2 channels. Because U-Net is usually used for image segmentation (classification), it was necessary to remove its final layer (softmax). For the first encoder, 64 filters in convolution layers were used. For the remaining layers, the structure was adjusted according to the original intention of the U-Net creators [26]. The network used 3 × 3 convolutions and 2 × 2 maximum pooling. The network had over 31 million parameters (learnables) and 60 layers.

During training, a loss function value for the mini batch was computed using the form derived by Nix & Weigend [28] and Bishop [29] to enable the neural network to account for heteroscedasticity. The derivation of this loss function starts from defining the likelihood $\mathcal{L}$ as

$$\mathcal{L} = \prod_l p(\boldsymbol{x}^l, \boldsymbol{T}^l) = \prod_l p(\boldsymbol{T}^l|\boldsymbol{x}^l)p(\boldsymbol{x}^l), \quad (1)$$

where $l$ is an example number in the mini batch, $\boldsymbol{T}^l$ are target patches, and $\boldsymbol{x}^l$ are input patches. In Equation (1), it is assumed that each data point $(\boldsymbol{x}^l, \boldsymbol{T}^l)$ is drawn independently from the same distribution, which justifies multiplying the probabilities. Next, for computational convenience, the loss function $E$ is constructed as the negative natural logarithm of the likelihood; minimizing this loss function is equivalent to maximizing the likelihood

$$E = -\ln\mathcal{L} = -\sum_l \ln p(\boldsymbol{T}^l|\boldsymbol{x}^l) - \sum_l \ln p(\boldsymbol{x}^l). \quad (2)$$

In a subsequent step, the second term resulting from the logarithm of the probability of the input data is dropped because it does not depend on the neural network weights, leading to a loss function of the form

$$E = -\sum_l \ln p(\boldsymbol{T}^l|\boldsymbol{x}^l). \quad (3)$$

To account for input-dependent variance, the conditional distribution of the target variables is written (assuming a normal distribution):

$$p\left(T_{(r,c)}|\boldsymbol{x}^l\right) = \frac{1}{\sqrt{2\pi}\sigma_{(r,c)}(\boldsymbol{x}^l)}\exp\left(-\frac{\left(y_{(r,c,1)}(x^l)-T_{(r,c)}\right)^2}{2\sigma^2_{(r,c)}(x^l)}\right), \tag{4}$$

where $r$ and $c$ represent the row and column numbers of the output (or target), respectively, $y_{(r,c,1)}(\boldsymbol{x}^l)$ are the values predicted by the output units in the first channel, and $\sigma_{(r,c)}(\boldsymbol{x}^l)$ is the input-dependent standard deviation. Combining this definition of the conditional distribution with Equation (3) leads to a loss function of the form

$$E = \sum_{l=1}^{L}\sum_{r=1}^{R}\sum_{c=1}^{C}\left(\ln\sigma_{(r,c)}\,(\boldsymbol{x}^l) + \frac{(y_{(r,c,1)}(x^l)-T^l_{(r,c)})^2}{2\sigma^2_{(r,c)}(x^l)}\right). \tag{5}$$

By utilizing the properties of the logarithms and moving a constant from the second factor before the fraction, Equation (5) is transformed into the form

$$E = \sum_{l=1}^{L}\sum_{r=1}^{R}\sum_{c=1}^{C}\left(\frac{1}{2}\ln\sigma^2_{(r,c)}(\boldsymbol{x}^l) + \frac{1}{2}\frac{(y_{(r,c,1)}(x^l)-T^l_{(r,c)})^2}{\sigma^2_{(r,c)}(x^l)}\right). \tag{6}$$

To enable the neural network to account for heteroscedasticity, a second output channel $y_{(r,c,2)}(\boldsymbol{x}^l)$ is introduced to model $\ln\sigma^2_{(r,c)}(\boldsymbol{x}^l)$. Modeling the logarithm of the variance instead of the variance itself is advantageous because it helps avoid potential issues related to the interpretation of negative values that could arise at the output of the second channel if it were directly predicting the variance. Finally, for convenience, a normalizing constant $\frac{1}{L}\cdot\frac{1}{R}\cdot\frac{1}{C}$ is added before the sums, yielding the final, implemented version of the loss function:

$$E = \frac{1}{L}\cdot\frac{1}{R}\cdot\frac{1}{C}\sum_{l=1}^{L}\sum_{r=1}^{R}\sum_{c=1}^{C}\left(\frac{1}{2}y_{(r,c,2)}(\boldsymbol{x}^l) + \frac{1}{2}\frac{(y_{(r,c,1)}(x^l)-T^l_{(r,c)})^2}{\exp(y_{(r,c,2)}(x^l))}\right). \tag{7}$$

It can be noted that model trained with the use of the loss function described by Equation (7) will aim to jointly estimate the conditional average value in the first output channel $y_{(r,c,1)}(\boldsymbol{x}^l)$ and the natural logarithm of the conditional variance in the second output channel $y_{(r,c,2)}(\boldsymbol{x}^l)$. It should be noted that the last term in Equation (7) minimizes relative errors, not absolute errors.. The derivation of this loss function assumes a normal conditional error distribution. While this assumption is difficult to verify, it is justified in this application since uncertainty is used solely as a measure for classifying cells within the uniformed point cloud.

During prediction (after training), determining the uncertainty $u^l_{(r,c)}$ in the correction estimates requires an explicit transformation:

$$u^l_{(r,c)} = \sqrt{\exp(y_{(r,c,2)}(\boldsymbol{x}^l))}. \tag{8}$$

During training, one random patch was cropped from each of approximately 4100 training images in every epoch. These images were generated from training region data, including both inputs and reference outputs. Typically, the full images measured 100 m × 100 m (2000 px × 2000 px), although some were smaller near the edges of the region. If a randomly selected patch lacked complete input or output data, a new patch was drawn. To augment the training dataset, random rotations were applied.

Validation images were cropped systematically on a 15 m × 15 m grid from approximately 1400 images (100 m × 100 m) derived from validation region data, resulting in about 31,000 validation examples. The goal of using a fixed set of validation examples—rather than dynamically cropping them during training—was to maintain a consistent and repeatable baseline for validation checks, ensuring that results were not affected by random cropping or rotation.

The root mean square propagation algorithm [29] was used for training the network, with its parameters selected through trial and error. After extensive experimentation, the following final parameter values were adopted: initial learn rate: 0.0001; learn rate schedule: piecewise; learn rate drop factor: 0.80; learn rate drop period: 10; L2 regularization factor: 0.0002; max epochs: 400; mini batch size: 128. Training examples were shuffled in every epoch. Validation data was used for early stopping, with a validation frequency of once per 50 iterations and validation patience of 20 checks. The training was stopped after 328 epochs. The end of training was caused by reaching the minimum value of the loss function (−1.80).

#### 2.1.4. Vertical Displacement Determination

In the proposed algorithm, vertical displacements are determined at selected nodes (Figure 1), with their positions configurable as needed. The processing is conducted independently for each node. The first step involves predicting elevation corrections and the logarithms of their variances (enabling the estimation of correction uncertainties) for the region surrounding each node. These values are derived from the developed U-Net model that takes as input a 10.20 m × 10.20 m area (comprising 204 × 204 cells) centered on the node. As a result, corrections and uncertainties are determined for all cells within the *uni* that fall within the network's output region. For a single node, this corresponds to a 1 m × 1 m square (20 × 20 cells). Thus, for each node where vertical displacement is to be estimated, a total of 400 corrections and their corresponding 400 uncertainties are computed for each measurement series.

Corrected elevations are computed only for cells where the uncertainty of corrections is below a predefined threshold (ulim) as differences between elevation in *uni* and the corresponding correction. All other cells (with uncertainty above ulim) are excluded from the calculations. Additionally, cells where the absolute correction values exceed the threshold set at 5 m are also discarded. This exclusion is justified by the fact that training data omits cells with corrections exceeding this limit, which in turn is based on the maximum vegetation elevation for which effective correction estimation is feasible.

The corrected elevations of cells from *uni* are used to determine the corrected elevation of each node. This elevation is computed as the median of all corrected cell elevations within a 1 m × 1 m square (the area covered by the network outputs), centered on the node where subsidence is being estimated. Using the median reduces the impact of errors, including outliers, and mitigates the influence of high-frequency (in the spatial domain) vertical ground surface displacements caused by agricultural activities (e.g., plowing).

The vertical displacement of a node is determined as the difference between the corrected node elevations from the two measurement series. Nodes are excluded from calculations if in one or both series, 50% or more of the cells within the 1 m × 1 m square (centered on the node) in the *uni* variable lack elevation data. This safeguard ensures that areas with a low data density, where extensive interpolation is required to fill gaps, are removed from the analysis.

Vertical displacements were determined on a regular 10 m × 10 m grid. Considering the accuracy of the subsidence determination and the nature of the phenomenon—specifically, terrain subsidence caused by underground mining operations, for which this algorithm was developed—such density appears sufficient. For all analyzed time intervals, calculations were performed using three different values of ulim: 15 cm, 20 cm, and 25 cm.

### *2.2. Vertical Displacement Determination Using Ground Filtering Algorithms*

The accuracy of vertical displacements determined using the neural network was compared with that achieved using selected ground filtering algorithms. The following widely used algorithms were tested in this context: ATIN [8], CSF [17], and SMRF [15]. In all cases, subsidence was determined as the difference between the median elevations of points classified as ground within a 1 m × 1 m square, centered on the node for which subsidence was being estimated. To ensure full comparability, subsidence was computed on the same 10 m × 10 m grid as in the U-Net-based algorithm.

To complement the comparison, subsidence was also estimated without applying any point cloud filtering (w/o filtering). In this case, subsidence was computed as the difference between median elevations, following the same approach described earlier, but under the assumption that all points were classified as ground. This approach was intended to establish a baseline against which other methods could be compared. This is justified by the fact that nearly every analyzed dataset included areas entirely devoid of vegetation, as well as regions with vegetation of varying height and density. Computing subsidence without filtering allows for a straightforward visual assessment of where filtering improves the results.

The tested ground filtering algorithms require setting parameters. The goal was to find a parameter set that allowed for subsidence determination over the largest possible area while minimizing errors, particularly outliers in the determined values. The optimization process began with a few selected parameter sets, which were gradually expanded as needed until further optimization no longer yielded significant improvements. In every case, the same parameter sets were applied to both point clouds used for subsidence estimation. This approach is justified because the datasets were selected to ensure that in each pair of measurement series, one was collected when vegetation influence was negligible. For such a low-vegetation series, parameter selection had minimal impact, meaning that the final accuracy of the determined vertical displacements depended on the filtering results of the series acquired under conditions of significant vegetation influence.

For the ATIN algorithm, the implementation available in Agisoft Metashape Professional ver. 2.0.3 was used. The following fixed parameter values were applied: cell size: 50 m; erosion radius: 0 m; return number: “Any return”. The two key filtering parameters, max angle (ma) and max distance (md), were tested in the following baseline configurations: ma: 10°, 15°, and 25°; md: 5 cm and 15 cm.

The CSF algorithm was used in its Matlab implementation (ver. 1.2.3), available on mathworks.com. The following parameter settings were tested: rigidness (rig): 1, 2, and 3; isSmooth: true; cloth resolution: 0.5 m; classification threshold: 5 cm; iterations: 600; and time step: 0.65. Alternative parameter sets (e.g., increasing cloth resolution, setting isSmooth to false, or increasing the classification threshold to 10 cm, 15 cm, or 50 cm) were tested on a few datasets, but they did not improve results. The final reported results are based only on the three sets of parameters, varying the rig value (rig: 1, 2, and 3), tested across all analyzed observation series pairs.

The SMRF algorithm was tested using its Matlab 2022b implementation. This algorithm requires the following parameters: grid resolution: 0.5 m; max window radius: 18 m; slope threshold (st): 0.001 to 0.15; elevation threshold: 5 cm; elevation scale: 1.25. The algorithm was also tested with a smaller grid resolution, but this significantly increased the processing time while having a negligible impact on accuracy. As a result, grid densification was abandoned.

### *2.3. Assessment and Ranking of Results*

Three evaluation metrics were used to assess the results. When applied together, these metrics provide a comprehensive representation of the quality of the results: the RMSE (the lower, the better); the percentage of nodes with determined vertical displacements relative to all nodes for which the vertical displacements could be determined (the higher, the better); and the percentage of outliers among nodes with determined vertical displacements (the lower, the better).

The RMSE was calculated according to the equation

$$\mathrm{RMSE} = \sqrt{\frac{\sum_{i=1}^{n} \delta_i^2}{n}}, \tag{9}$$

where $\delta_i$ are errors of determined vertical displacements, and $n$ is a number of determined vertical displacements. This statistic was calculated taking into account all nodes for which the vertical displacements were determined with the exclusion of nodes for which the vertical displacements were determined with gross errors. Based on previous experiences and the expected results (error levels), it was assumed that a vertical displacement determination would be considered an outlier if its absolute error was greater than 45 cm.

Taking into account only one of the proposed quality indicators in the evaluation algorithms for subsidence determination would lead to biased results of the comparison of performance. Generally, a change in the algorithm parameter values resulting in an increase in the number of nodes for which displacements are determined is usually (for a given case) associated with an increase in the values of the other two quality indicators (RMSE, percentage of outliers). For this reason, in order to perform the algorithm performance ranking, it was necessary to take into account all of these parameters together, so that the highest rated algorithm would guarantee the best proportion between quality measures. Result ranking was conducted using the technique for order of preference by similarity to ideal solution (TOPSIS) methodology [30]. This technique is used to rank objects described by multidimensional criteria through the determination of their distance in multidimensional space from the potentially most favorable and least favorable variants. These variants are determined by identifying, for each criterion, the highest and lowest rated values. As a result, two (not necessarily existing in the dataset) variants are created with the best and the worst properties. The distance, in the multidimensional space, from both of them is a measure of the quality of vertical displacement determination, according to the formula

$$\mathrm{TIndex}_j = \frac{DistWORST_j}{DistWORST_j + DistBEST_j}, \tag{10}$$

where $j$ is a variant number, $\mathrm{TIndex}_j$ is a ranking coefficient, $DistWORST_j$ is the Euclidean distance in multidimensional space from the worst variant, and $DistBEST_j$ is the Euclidean distance in multidimensional space from the most favorable variant. Due to the different ranges of variability among the three evaluated characteristics, normalization is crucial for constructing a ranking. No weight matrix was used, which means that all three features had the same impact on the TOPSIS score. Normalization was performed using the L2 norm (vector normalization). Importantly, the w/o filtering solution was also included in both the normalization and ranking processes. In the final result tables, only the best-performing configuration (i.e., the one yielding the highest TIndex) was reported for each method.

### *2.4. Datasets*

All study sites are located in southern Poland, and their surface topography is relatively simple. The sites are generally flat or consist of gentle hillslopes with low levels of urbanization, typical of rural landscapes.

Due to the presence of buildings and forested areas, not all parts of the study sites were used in the research. The regions utilized for training, validation, and testing of the neural networks are marked with colored outlines in Figure 4 (green for training, yellow for validation, and red for testing). The validation and test regions were also analyzed for vertical displacement accuracy using ground filtering methods. Detailed descriptions of each dataset are provided in the following subsections.

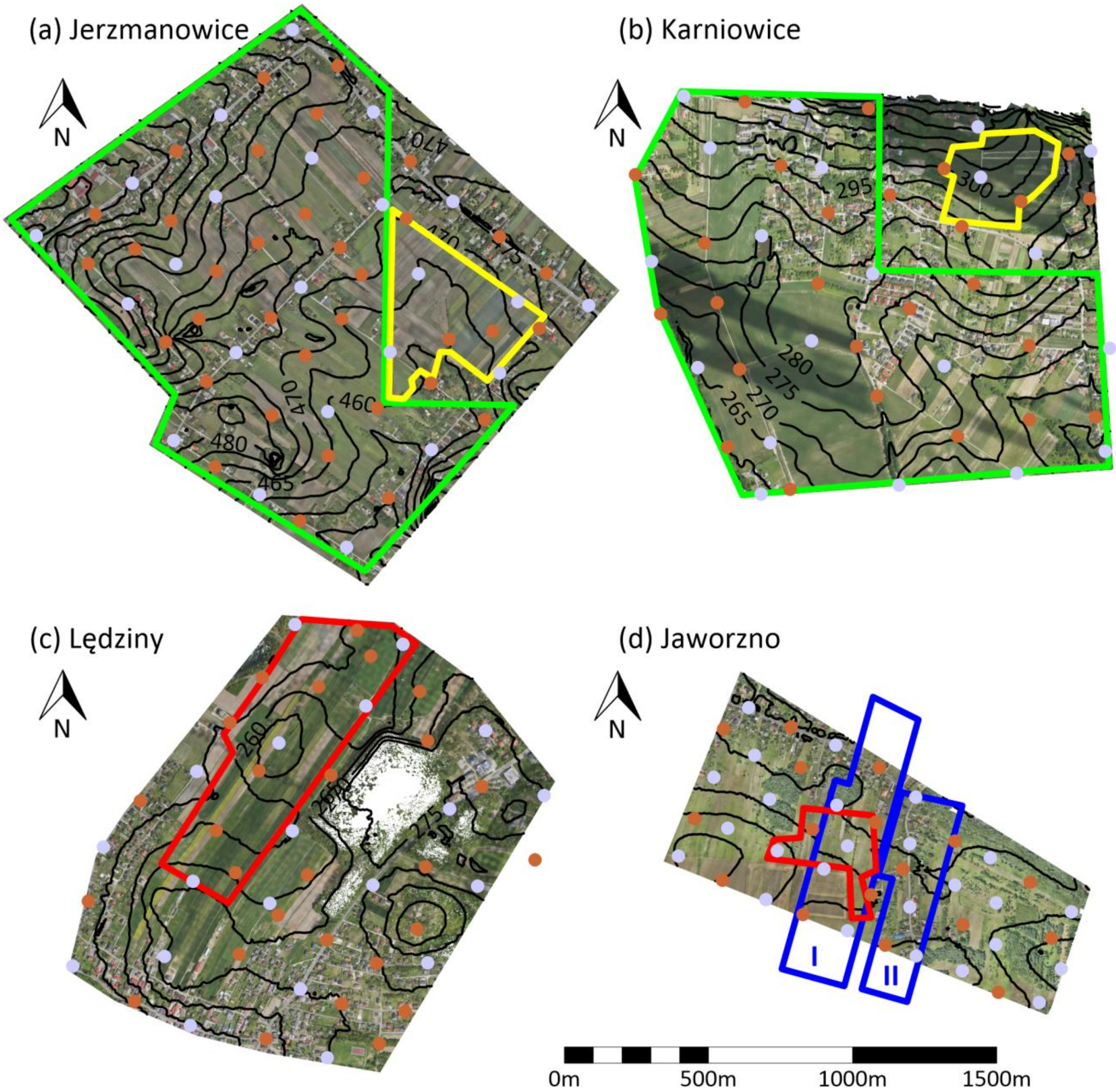

**Figure 4.** Visualization of study sites: (**a**) Jerzmanowice, (**b**) Karniowice, (**c**) Lędziny, (**d**) Jaworzno. Training regions are outlined in green, validation regions are outlined in yellow, and test regions are outlined in red. The underground mining longwall contours are marked in blue. The check points are marked as light purple dots and ground control points as light brown dots.

In the first three study sites (Figure 4a–c), within the boundaries of analyzed regions (green, yellow and red outlines), there are no vertical ground displacements. The mining operation contours in Jaworzno, representing the mining activity conducted between measurement series, are outlined in blue in Figure 4d. This mining activity has induced ground surface displacements.

### 2.4.1. Jerzmanowice

The Jerzmanowice dataset (Figure 4a) consists of 12 UAV-photogrammetry series covering an area of approximately 220 ha in Jerzmanowice (19°45′N, 50°13′E), located in the Lesser Poland Voivodeship. The topography is relatively simple, with local slopes reaching up to 17 degrees, though most of the area has a slope of 5 degrees or less. The land cover primarily consists of fields and meadows with scattered trees. Buildings, mainly single-family houses and farm structures, are concentrated along roads, which run near the borders of the site and cross its central section. The vegetation on cultivated land is diverse and includes grasses and cereals, including maize.

The observations were conducted using a DJI (Shenzhen, China) Zenmuse P1 camera with a 35 mm f/2.8 lens, mounted on a DJI Matrice 300 RTK multirotor platform. Photogrammetric missions were carried out in follow terrain mode, ensuring a uniform GSD of 10 mm across the entire site. The flights were executed with an 80% forward and 60% side overlap. Measurements were referenced to ground control points (GCPs) at a density of 1 GCP per 10 ha. The GCP positions were determined using GNSS techniques: static GNSS along roads and GNSS RTK with a dedicated base receiver for other locations. Additionally, 36 control points (CPs) were evenly distributed and measured using GNSS RTK for accuracy verification. The GCPs and CPs coordinates were measured once before the first series and then subsequently verified before each series. An accuracy assessment based on GCP and CP error analysis showed a reprojection accuracy of approximately 11 mm in the horizontal plane and 17 mm in the vertical plane. Dense point clouds were generated under the assumption that spatial coordinates were computed for each 2 × 2 pixel block in the images.

The Jerzmanowice study site was divided into a training region for neural networks (approximately 160 ha) and a validation region (approximately 20 ha). The dates of measurement series used for training, validation, and testing of the neural networks are listed in Table 1. Since the Jerzmanowice site does not undergo any displacements, the reference subsidence value was set to zero. Any deviation from this value can be considered an error in the estimated vertical displacement values.

The accuracy of vertical displacement determination using both ground filtering algorithms and neural networks was evaluated based on the following time intervals: 2022.03.–2022.06., 2022.03.–2022.09. This analysis was conducted for the validation region (marked in yellow in Figure 4a). During the 2022.03. series (early spring), the influence of vegetation on ground elevation measurements was negligible. However, in the 2022.06. (summer, peak vegetation season) and 2022.09. (autumn) series, significant vegetation growth was present in some parts of the validation region, particularly in corn-covered fields.

**Table 1.** Summary of datasets and their role in U-Net training and evaluation.

| Dataset | Measurement Series Dates | Application |
|---|---|---|
| Jerzmanowice | 2021.10., 2021.11., 2021.12., 2022.02., 2022.03., 2022.04., 2022.05., 2022.06., 2022.07., 2022.08., 2022.09., 2023.01. | Training and validation |
| Karniowice | 2021.11., 2022.03., 2022.04., 2022.05., 2022.06., 2022.08., 2022.09., 2022.10., 2022.12., 2023.02. | Training and validation |
| Lędziny | 2021.11., 2022.05., 2022.08., 2022.10. | Test |
| Jaworzno | 2019.07., 2020.11., 2021.11. | Test |

Additionally, the Jerzmanowice dataset included a cleaned ALS point cloud from 2019.10., with a density of approximately 4 points/m$^2$ and GNSS RTK measurement results, referenced to a dedicated base station, conducted on agricultural land in a 50 m × 50 m grid in 2021.12.

2.4.2. Karniowice

The Karniowice dataset (Figure 4b) consists of 10 UAV-photogrammetry series, covering an area of approximately 200 ha in Karniowice (19°47′N, 50°09′E)), Lesser Poland Voivodeship. The terrain slope reaches a maximum of about 6 degrees, though it does not exceed 3 degrees in most of the site. The land cover primarily consists of fields and meadows with scattered trees. Buildings are mainly single-family houses and farm structures.

In the northwestern part, there are multi-story residential buildings. In the eastern section, there are industrial halls with adjacent storage areas and parking lots. The vegetation on cultivated land varies and includes grasses, cereals, and maize, with a significant portion of the western area covered by cornfields.

The photogrammetric observations were conducted using a Sony RX1R II camera (Sony Corporation, Tokyo, Japan) mounted on a FlyTech (Krakow, Poland) Birdie fixed-wing platform equipped with a GNSS RTK module. Photogrammetric missions were carried out in follow terrain mode to maintain a consistent GSD of 10 mm across the entire site. The missions featured 50% forward and 70% side overlap, employing cross-flight patterns. Measurements were referenced to GCPs with a density of approximately one point per 10 hectares, along with 26 evenly distributed CPs. The positions of the GCPs and CPs were determined using GNSS RTK in reference to a dedicated base station. The positions of GCPs and CPs were measured independently in each series. Based on the error analysis for GCPs and CPs, the measurement data showed a reprojection accuracy of 17 mm in the horizontal plane and 25 mm in the vertical plane.

The Karniowice study site was divided into a training region for neural networks (approximately 150 ha) and a validation region (about 10 ha). The dates of measurement series used for training, validation, and testing of the U-Net are listed in Table 1. Since this area does not experience displacements, the reference subsidence value was set to zero. Any deviations from this value are considered errors in the determined subsidence values.

The accuracy of vertical displacement determination using both ground filtering algorithms and neural networks was evaluated for the following time intervals: 2022.03.–2022.06., 2022.03.–2022.09. This analysis was conducted for the validation region (marked in yellow in Figure 4b). During the 2022.03. series (early spring), the influence of vegetation on elevation determination was negligible. However, in the 2022.06 and 2022.09. series, significant vegetation growth was present in parts of the validation area, particularly in corn and cereal fields.

Additionally, the Karniowice dataset included a cleaned ALS point cloud from 2013.04., with a density of approximately 4 points/m$^2$, and GNSS RTK measurements, referenced to a dedicated base station, conducted on agricultural land in a 10 m × 10 m grid in 2021.11.

#### 2.4.3. Lędziny

The Lędziny dataset (Figure 4c) consists of four UAV-photogrammetry series (Table 1), covering an area of approximately 170 ha in Lędziny (19°07′N, 50°08′E), Silesian Voivodeship. The site is located within the Upper Silesian Coal Basin (USCB) and has terrain characteristics representative of this mining region. The maximum terrain slope is around 5 degrees, but it is 3 degrees or less for most of the site. The western part is mainly covered by fields and meadows, the eastern and southern sections are occupied by buildings, and a forest covers the central part of the site.

The observations were conducted using a Sony RX1R II camera, mounted on a FlyTech Birdie fixed-wing platform equipped with a GNSS RTK module. Photogrammetric missions were performed in follow terrain mode, ensuring a uniform GSD of 10 mm across the entire site. Flights were carried out with 50% forward overlap and 70% side overlap, using cross-flight patterns. Survey measurements were referenced to 20 GCPs at a density of 1 point per 8 ha, along with 27 evenly distributed CPs. The positions of the GCPs and CPs were determined using GNSS RTK, referenced to a dedicated base station. The positions of GCPs and CPs were measured independently in each series. An assessment based on GCP and CP error analysis indicated a reprojection accuracy of 15 mm in the horizontal plane and 20 mm in the vertical plane.

For testing, only the western part of the surveyed area, covering approximately 40 ha, was used. This region consists exclusively of fields and meadows. The remaining area was excluded from testing because it was either covered by buildings or located within the zone of influence of underground hard coal mining operations. The tested fragment was not affected by mining activities, meaning no surface displacements occurred in the analyzed period. As a result, the reference subsidence value was set to zero, and any deviation from this value was interpreted as an error in vertical displacement determination.

The accuracy of vertical displacement determination using both ground filtering algorithms and U-Net was evaluated for the following time intervals: 2021.11.–2022.05., 2021.11.–2022.08., and 2021.11–2022.10. This analysis was conducted for the test region, marked in red in Figure 4c. During the 2021.11. series (late autumn), the influence of vegetation on elevation measurements was negligible. In the other series, the influence of vegetation varied across the test region. It was significant in small sections of the region covered by cornfields and noticeable in the remaining part, primarily in meadows.

#### 2.4.4. Jaworzno

The Jaworzno dataset (Figure 4d) consists of three UAV-photogrammetry series, covering an area of approximately 80 ha in Jaworzno (19°20′N, 50°11′E), Silesian Voivodeship. The site is located within the USCB, and its topography is representative of this mining region. The maximum terrain slope is around 3 degrees, though it is even lower for most of the site. The western part is primarily covered by fields and meadows; the eastern part is mainly forested. The central and northern sections contain buildings, mostly single-family houses and farm structures.

The site was subject to the influence of underground hard coal mining operations. During the analyzed intervals, ongoing mining activity included the first and second longwalls in a coal seam located approximately 650 m below the surface. The mining height was approximately 2.8 m, and extraction was conducted with caving. In Figure 4d, the blue outlines mark the longwalls that were excavated between the measurement series. The mining sequence was as follows: the first longwall ("I") was mined from 2019.07. to 2020.11., and then the second longwall ("II") was mined from 2020.11. to 2021.11.

Observations in the 2019.07. series were made using a DJI S1000 platform (multirotor, without GNSS RTK module) with a Sony Alfa A7R camera (Sony, Tokyo, Japan) equipped with a Sony Zeiss Sonnar T* FE 35 mm F2.8 ZA lens. Observations in the other two series were made using a FlyTech Birdie equipped with a GNSS RTK module and a Sony RX1R II camera with a Carl Zeiss (Oberkochen, Germany) Sonnar T* 35 mm F2.0 lens. The photogrammetric missions were carried out in follow terrain mode so that the GSD for the entire site was at a uniform level of 10 mm, with forward overlap of 80% and side overlap of 60% for the DJI S1000 platform and forward overlap of 50% and side overlap of 70% using a cross-flight pattern for the FlyTech Birdie platform. The measurements were referenced to 25 GCPs (1 point per 3 ha) evenly distributed over the site. In addition, there were 17 CPs evenly distributed over the site to estimate the accuracy. The position of the GCPs and CPs was determined using GNSS RTK techniques in relation to the base station located nearby. The positions of GCPs and CPs were measured independently in each series. Based on the GCP and CP analysis, the reprojection error was estimated at 15 mm in the horizontal plane and 25 mm in the vertical plane.

The ground elevations in the test region were additionally measured in each observation series using GNSS RTK, referenced to a dedicated base station. These measurements were conducted on a 10 m × 10 m grid, and based on this data, interpolation using kriging was performed to model the terrain surface. To avoid large interpolation errors, the interpolation was applied only in fragments where the distance from grid points did not exceed 20 m. The differences between ground surface models created for each analyzed measurement series served as the reference subsidence model, against which errors for the analyzed ground filtering methods and U-Net were evaluated.

The accuracy of vertical displacement determination using both ground filtering algorithms and U-Net was evaluated for the following time intervals: 2019.07.–2020.11. (corresponding to the extraction of longwall "I" outlined in Figure 4d) and 2019.07.–2021.11. (corresponding to the extraction of longwalls "I" and "II" outlined in Figure 4d). The accuracy analysis was conducted for the test region, marked in red in Figure 4d. In the 2020.11. and 2021.11. series (late autumn), the influence of vegetation on terrain elevation determination was small. During the 2019.07. series, vegetation was at its peak just before the grain harvest, and the grass in the meadows was very tall and dense. However, due to the necessity of conducting GNSS RTK reference measurements, which could negatively impact crop yields, the evaluated fragment does not include grain fields but only meadows. It is important to note that the reference subsidence models derived from the GNSS RTK measurement and interpolation are not error-free. Cross-validation estimates using 20 folds indicate that the RMSE of the reference subsidence model is approximately 10–15 cm.

## 3. Results

The quality assessment and comparison were conducted for the validation regions of the Jerzmanowice and Karniowice datasets and the test regions of the Lędziny and Jaworzno datasets. The training regions were excluded, as they would have favored the U-Net-based algorithm and would not have accurately reflected its actual generalization capabilities.

### *3.1. Results for Validation Datasets*

Table 2 presents a summary of key characteristics and TIndex values for the validation regions of the Jerzmanowice and Karniowice datasets in the 2022.03–2022.06 and 2022.03–2022.09 intervals. According to this ranking method, the best results were achieved using a U-Net-based algorithm (three cases) and an ATIN-based algorithm (one case). The differences in TIndex values between these two algorithms were generally small, exceeding 0.05 only in extreme cases. In most cases, U-Net provided slightly higher accuracy in estimating subsidence. An algorithm based on SMRF performed reasonably well in most cases, except for the Karniowice dataset in the 2022.03–2022.06 interval. The CSF-based algorithm produced surprisingly weak results, comparable to the unfiltered (non-classified) point cloud approach.

**Table 2.** Summary of key characteristics for vertical displacement determination in validation regions. The methods are sorted according to their TIndex values.

| Dataset and Interval | Method and Parameters | RMSE [cm] | Percentage of Nodes with Determined Displacements [%] | Percentage of Outliers in Nodes with Determined Displacements [%] | TIndex |
|---|---|---|---|---|---|
| Jerzmanowice 2022.03.–2022.06. | ATIN, ma = 10 deg, md = 5 cm | 5.0 | 57.0 | 0.1 | 0.798 |
| | U-Net, ulim = 15 cm | 4.4 | 54.0 | 0.0 | 0.789 |
| | SMRF, st = 0.005 | 7.6 | 55.3 | 0.5 | 0.768 |
| | CSF, rig = 1 | 16.3 | 99.7 | 36.0 | 0.295 |
| | w/o filtering | 15.9 | 100.0 | 37.9 | 0.288 |
| Jerzmanowice 2022.03.–2022.09. | U-Net, ulim = 20 cm | 6.7 | 70.6 | 0.0 | 0.832 |
| | ATIN, ma = 10 deg, md = 5 cm | 9.1 | 66.7 | 0.0 | 0.778 |
| | SMRF, st = 0.01 | 9.2 | 67.2 | 0.5 | 0.775 |
| | CSF, rig = 1 | 12.8 | 99.6 | 30.5 | 0.276 |
| | w/o filtering | 12.9 | 100.0 | 31.6 | 0.270 |
| Karniowice 2022.03.–2022.06. | U-Net, ulim = 15 cm | 12.4 | 24.4 | 7.6 | 0.571 |
| | ATIN, ma = 10 deg, md = 5 cm | 8.3 | 22.3 | 16.6 | 0.570 |
| | w/o filtering | 19.5 | 93.3 | 71.7 | 0.471 |
| | CSF, rig = 2 | 18.7 | 88.4 | 70.3 | 0.468 |
| | SMRF, st = 0.01 | 15.4 | 50.5 | 53.9 | 0.431 |
| Karniowice 2022.03.–2022.09. | U-Net, ulim = 15 cm | 8.8 | 87.2 | 0.2 | 0.938 |
| | ATIN, ma = 10 deg, md = 5 cm | 12.0 | 87.8 | 0.1 | 0.894 |
| | SMRF, st = 0.01 | 12.8 | 71.1 | 0.4 | 0.831 |
| | CSF, rig = 3 | 14.3 | 95.5 | 8.7 | 0.302 |
| | w/o filtering | 15.3 | 100.0 | 11.4 | 0.205 |

For the Jerzmanowice dataset, in the 2022.03.–2022.06. interval, a graphical comparison of subsidence error distributions was performed. The errors were categorized into four groups: ≤15 cm (useful results), 15 cm–30 cm (limited usability), 30 cm–45 cm (generally unusable results), and >45 cm (outliers). According to this summary, the CSF filtering had almost no impact on subsidence error reduction, indicating its ineffectiveness in this scenario. For U-Net, the best results (highest TIndex value) were obtained with ulim set to 15 cm (Figure 5e). For comparison, Figure 5f contains results for ulim set to 20 cm. It can be noted that increasing the ulim threshold enabled subsidence estimation in areas with dense vegetation. However, the uncertainty of these additional estimations was noticeably higher than when using a lower ulim value. The detailed analyses of Figures 4 and 5 show that green dots (low-error subsidence determinations) for algorithms based on ATIN, U-Net, and SMRF in high-vegetation areas were generally associated with locally sparse vegetation or flattened vegetation due to machinery tracks, etc.

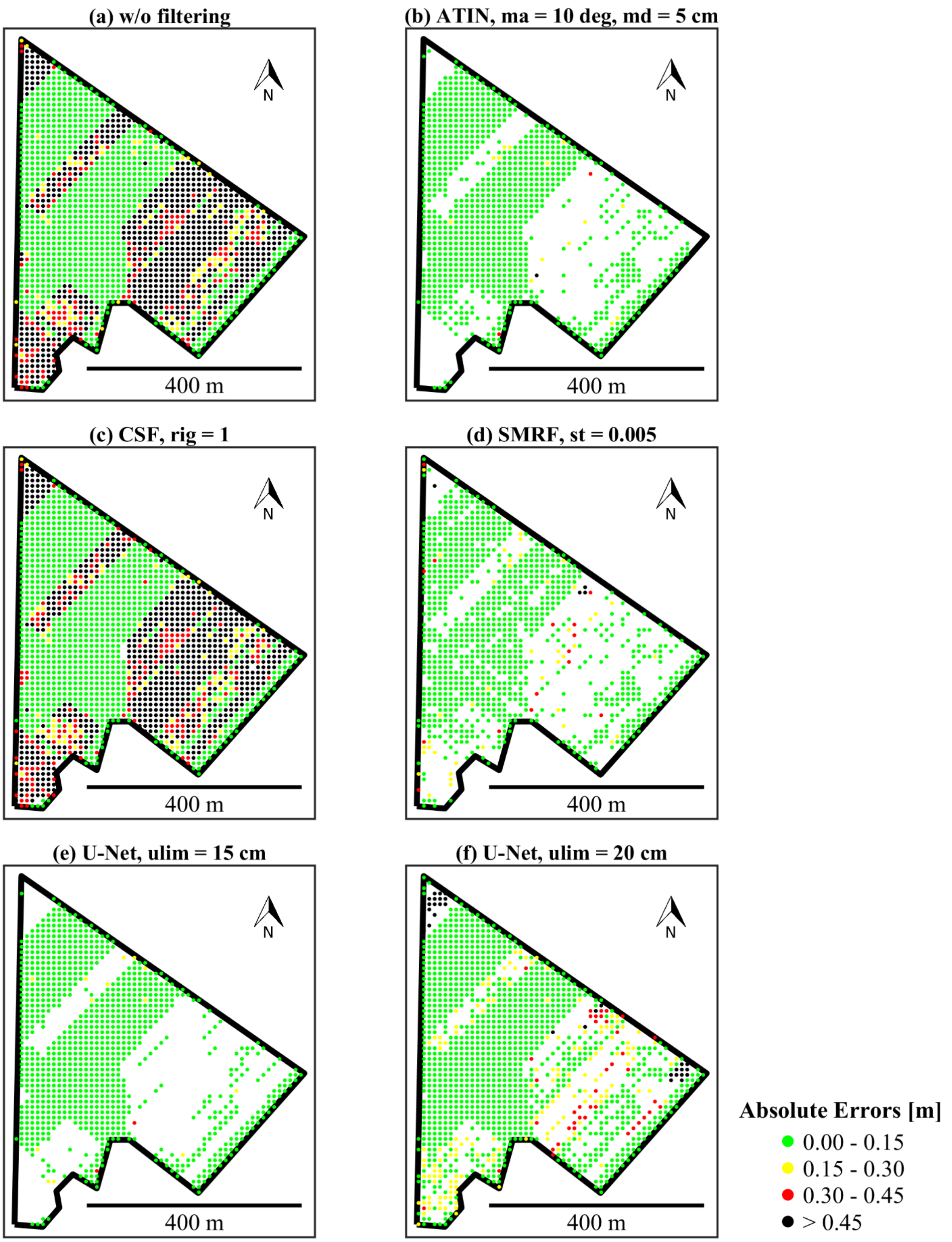

**Figure 5.** Summary of errors of the determined vertical displacements for the Jerzmanowice site in the interval 2022.03.–2022.06. for the validation region: (a) w/o filtering; (b) ATIN, ma = 10 deg, md = 5 cm; (c) CSF, rig = 1; (d) SMRF, st = 0.005; (e) U-Net, ulim = 15 cm; (f) U-Net, ulim = 20 cm.

Figure 6 presents the kernel-smoothed probability density function estimate of errors for the w/o filtering variant and the ATIN- and U-Net-based algorithms. The analysis of this figure, along with the accompanying statistics, provides insight into the expected error values and differences in error distributions among the selected algorithms. Both ATIN and U-Net effectively eliminated most large errors in vertical displacement determination. In this case, ATIN produced a mean error closer to zero. However, U-Net demonstrated a lower RMSE and a more compact error distribution, indicating greater accuracy.

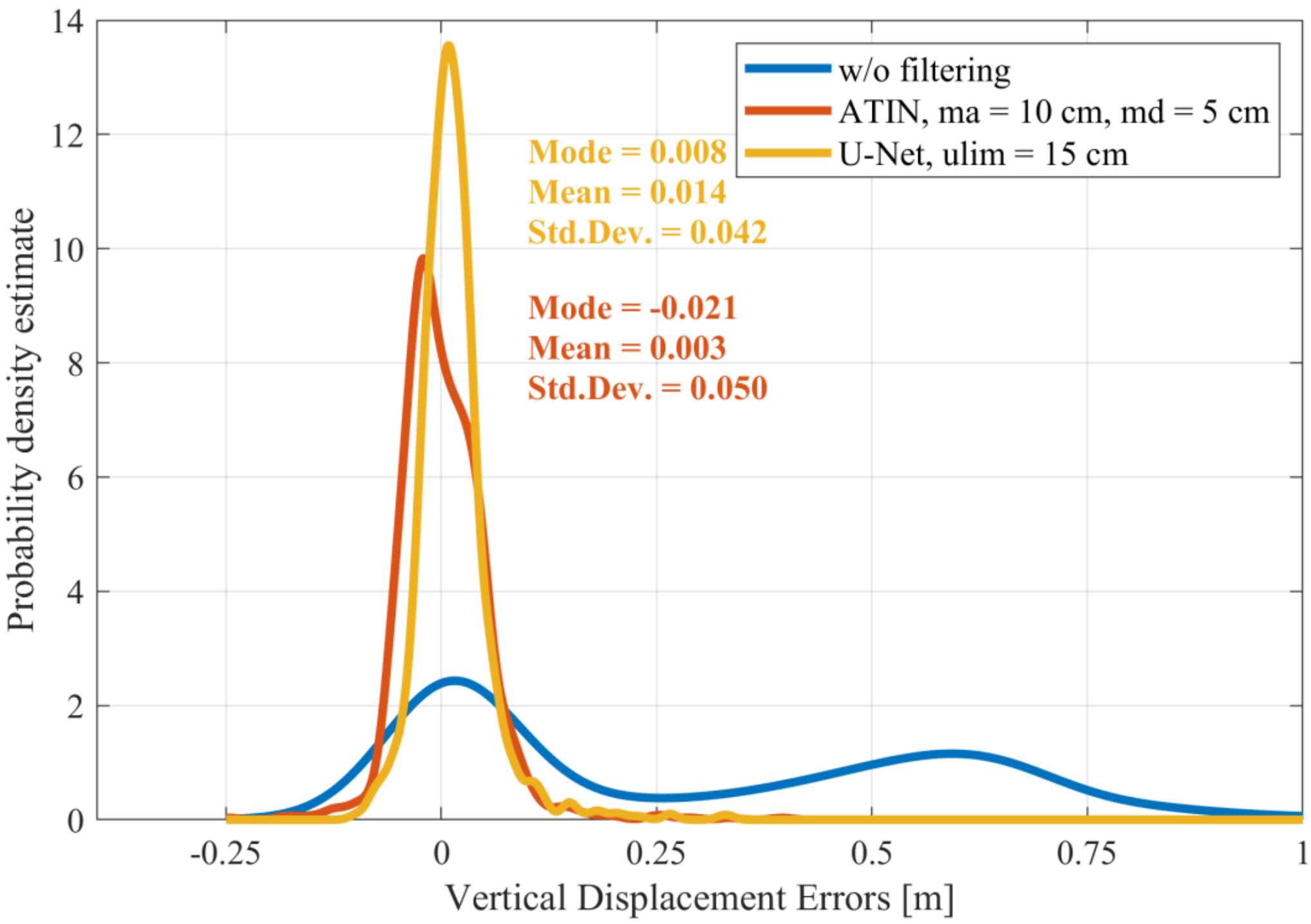

**Figure 6.** Probability density estimates of vertical displacement errors for the Jerzmanowice dataset (2022.03.–2022.06. interval, validation region). To ensure consistency with the statistics in Table 2, descriptive statistics were computed using only errors with absolute values below 0.45 m.

### *3.2. Results for Test Datasets*

The comparison results for the test datasets are even more insightful than those for the validation datasets. While the validation results may carry some bias—since early stopping and model selection were based on the loss function value on the validation dataset—this concern does not apply to the test dataset results. Additionally, the test regions are located directly within the USCB, making them fully representative in terms of both surface topography and vegetation coverage.

Table 3 presents the key characteristics and TIndex values for subsidence determination in the test datasets. Similarly to the validation datasets, the U-Net-based algorithm most frequently achieved the highest ranking (four times). In the fifth case (Jaworzno, 2019.07–2021.11), the TIndex value difference between U-Net and ATIN was only 0.0001. Given the uncertainty in the reference subsidence values, this difference can be considered negligible. As observed in the validation datasets, ATIN generally ranked high, but in two test cases (Lędziny, 2021.11–2022.08 and 2021.11–2022.10), the SMRF-based algorithm outperformed ATIN. Consistent with the validation dataset results, the CSF-based algorithm ranked low across all analyzed series. In one case (Lędziny, 2021.11–2022.05), it even performed worse than the w/o filtering approach.

**Table 3.** Summary of key characteristics for vertical displacement determination in test regions. The methods are sorted according to their TIndex values.

| Dataset and Interval | Method and Parameters | RMSE [cm] | Percentage of Nodes with Determined Displacements [%] | Percentage of Gross Errors in Nodes with Determined Displacements [%] | TIndex |
|---|---|---|---|---|---|
| Lędziny 2021.11.–2022.05. | U-Net, ulim = 15 cm | 9.7 | 64.7 | 0.7 | 0.789 |
| | ATIN, ma = 35 deg, md = 20 cm | 18.6 | 81.2 | 0.8 | 0.763 |
| | SMRF, st = 0.15 | 18.3 | 68.1 | 2.0 | 0.635 |
| | w/o filtering | 19.6 | 100.0 | 5.7 | 0.388 |
| | CSF, rig = 2 | 19.7 | 96.2 | 5.5 | 0.385 |
| Lędziny 2021.11.–2022.08. | U-Net, ulim = 20 cm | 6.1 | 80.5 | 0.2 | 0.885 |
| | SMRF, st = 0.01 | 7.7 | 77.3 | 0.3 | 0.832 |
| | ATIN, ma = 10 deg, md = 5 cm | 8.2 | 78.6 | 0.2 | 0.822 |
| | CSF, rig = 1 | 12.2 | 98.9 | 16.7 | 0.181 |
| | w/o filtering | 12.7 | 100.0 | 19.3 | 0.133 |
| Lędziny 2021.11.–2022.10. | U-Net, ulim = 25 cm | 5.9 | 91.5 | 0.1 | 0.949 |
| | SMRF, st = 0.01 | 7.2 | 84.6 | 0.1 | 0.884 |
| | ATIN, ma = 10 deg, md = 5 cm | 9.1 | 90.1 | 0.1 | 0.823 |
| | CSF, rig = 1 | 11.0 | 99.2 | 7.8 | 0.169 |
| | w/o filtering | 11.0 | 100.0 | 9.3 | 0.079 |
| Jaworzno 2019.07.–2020.11. | U-Net, ulim = 25 cm | 10.0 | 89.5 | 0.5 | 0.939 |
| | ATIN, ma = 10 deg, md = 15 cm | 11.3 | 89.1 | 0.3 | 0.924 |
| | SMRF, st = 0.03 | 12.7 | 90.3 | 0.0 | 0.904 |
| | CSF, rig = 3 | 17.2 | 96.2 | 6.8 | 0.474 |
| | w/o filtering | 18.2 | 99.8 | 13.2 | 0.125 |
| Jaworzno 2019.07.–2021.11. | ATIN, ma = 10 deg, md = 15 cm | 11.4 | 88.3 | 0.3 | 0.922 |
| | U-Net, ulim = 20 cm | 10.4 | 84.4 | 0.5 | 0.922 |
| | SMRF, st = 0.05 | 13.3 | 92.3 | 1.0 | 0.870 |
| | CSF, rig = 2 | 17.6 | 96.5 | 6.6 | 0.497 |
| | w/o filtering | 18.2 | 100.0 | 13.5 | 0.143 |

For the Lędziny dataset in the 2021.11.–2022.08. interval, a visualization of subsidence estimation errors was generated for all tested algorithms (Figure 7). As observed in the validation datasets, the CSF-based algorithm (Figure 7c) showed minimal impact from filtering, with results nearly identical to the unfiltered data (compared with Figure 7a). However, the differences between the ATIN-, SMRF-, and U-Net-based algorithms are particularly interesting. SMRF- (Figure 7d) and ATIN-based algorithms (Figure 7b) produced relatively similar results. In contrast, the U-Net-based algorithm (shown with a ulim value of 20 cm in Table 3 and Figure 7f) was able to effectively correct point cloud elevations in the northern part of the region. For comparison, results for U-Net with a ulim value of 15 cm are also shown (Figure 7e). A comparison of U-Net results for ulim values of 15 cm and 20 cm clearly indicates that the regions where SMRF and ATIN determined subsidence with relatively high errors also exhibited high uncertainty in correction values. However, higher correction uncertainty does not necessarily correspond to larger correction errors. Instead, it indicates a greater likelihood of larger errors rather than guaranteeing their presence.

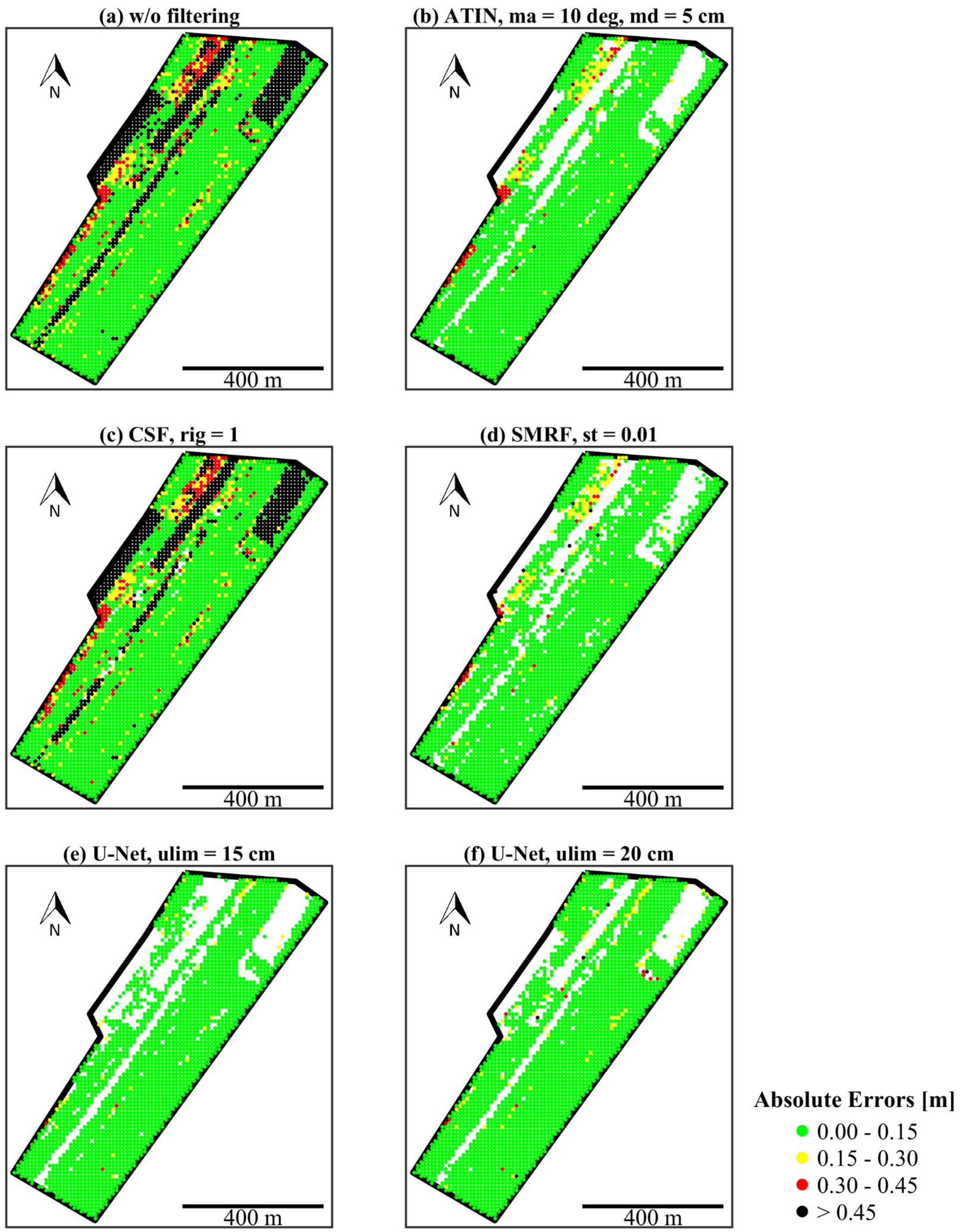

**Figure 7.** Summary of errors of the determined vertical displacements for the Lędziny in the interval 2021.11.–2022.08: (a) w/o filtering; (b) ATIN, ma = 10 deg, md = 5 cm; (c) CSF, rig = 1; (d) SMRF, st = 0.01; (e) U-Net, ulim = 15 cm; (f) U-Net, ulim = 20 cm.

Figure 8 presents the kernel-smoothed probability density function estimate of errors for the w/o filtering variant and the ATIN- and U-Net-based algorithms. Both ATIN and U-Net effectively eliminated most large errors in vertical displacement determination. In this case, both the mean error and error spread are smaller for the U-Net-based algorithm compared to ATIN.

**Figure 8.** Probability density estimates of vertical displacement errors for the Lędziny dataset (2021.11.–2022.08. interval). To ensure consistency with the statistics in Table 3, descriptive statistics were computed using only errors with absolute values below 0.45 m.

For the Jaworzno dataset, the U-Net was also used to determine subsidence across the entire area for the 2019.07–2021.11 interval, corresponding to underground hard coal mining operations in both longwalls (Figure 9). In this figure, vertical displacements outside the range of −2.2 m to 0.2 m were marked as outliers. This range was set with a 20 cm margin relative to the recorded reference displacement values. Comparing U-Net-based results with results of a variant w/o filtering clearly highlights where and to what extent elevation corrections influenced the interpretability of surface deformations. Large areas without U-Net-determined subsidence primarily correspond to forested regions (see Figure 4d) and agricultural fields just before harvest in 2019.07.

A surface deformation anomaly is clearly visible as a southeast-oriented linear feature in the central part of the figure, characterized by noticeably lower subsidence values compared to surrounding areas. This anomaly was previously described [1] and is most likely caused by subsurface layer heterogeneity. The U-Net-based subsidence map also reveals a clear asymmetry between the eastern and western flanks of the subsidence basin. These findings confirm that U-Net-derived results are not only sufficiently accurate for assessing overall surface displacement magnitudes but also detailed enough to support more in-depth analyses.

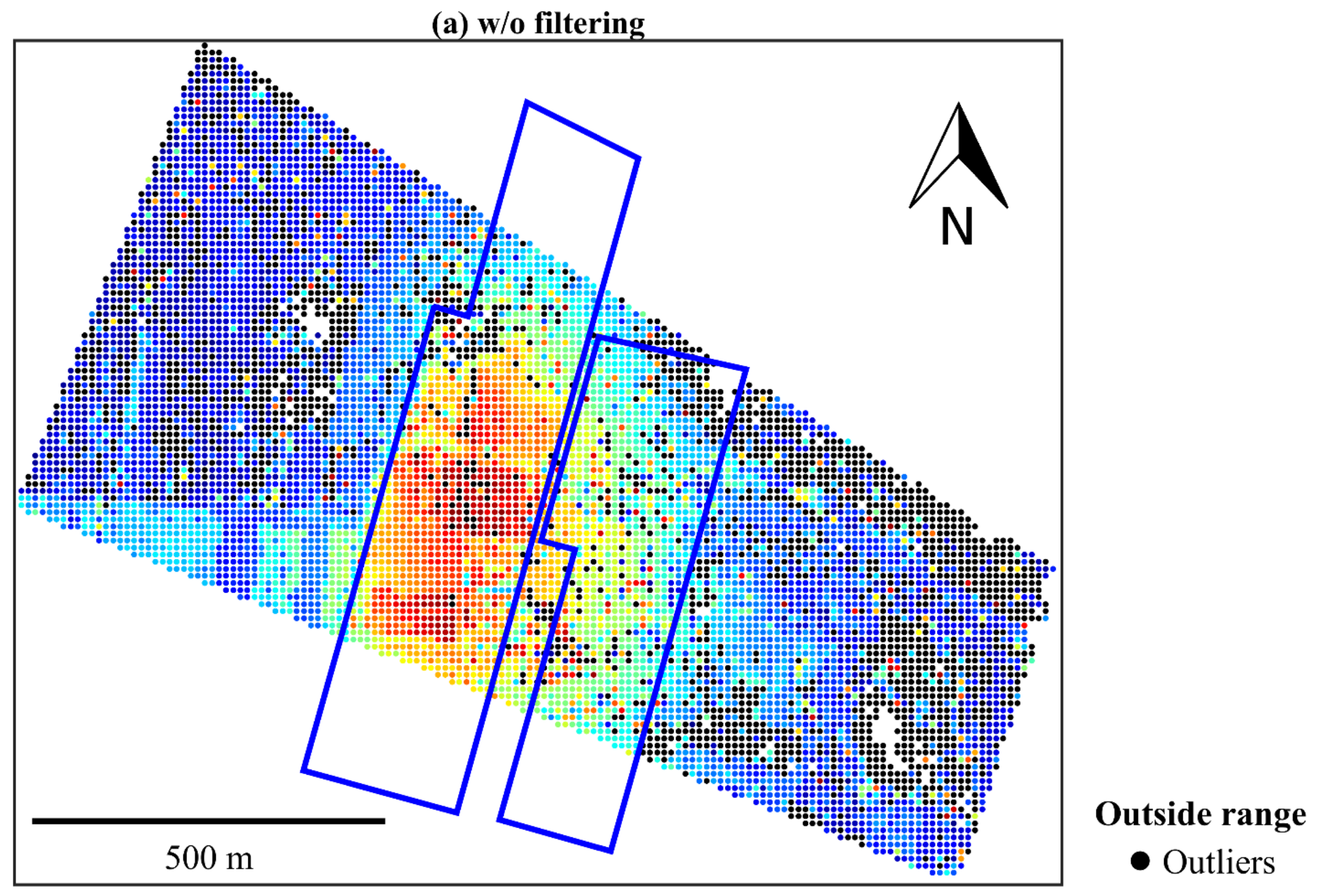

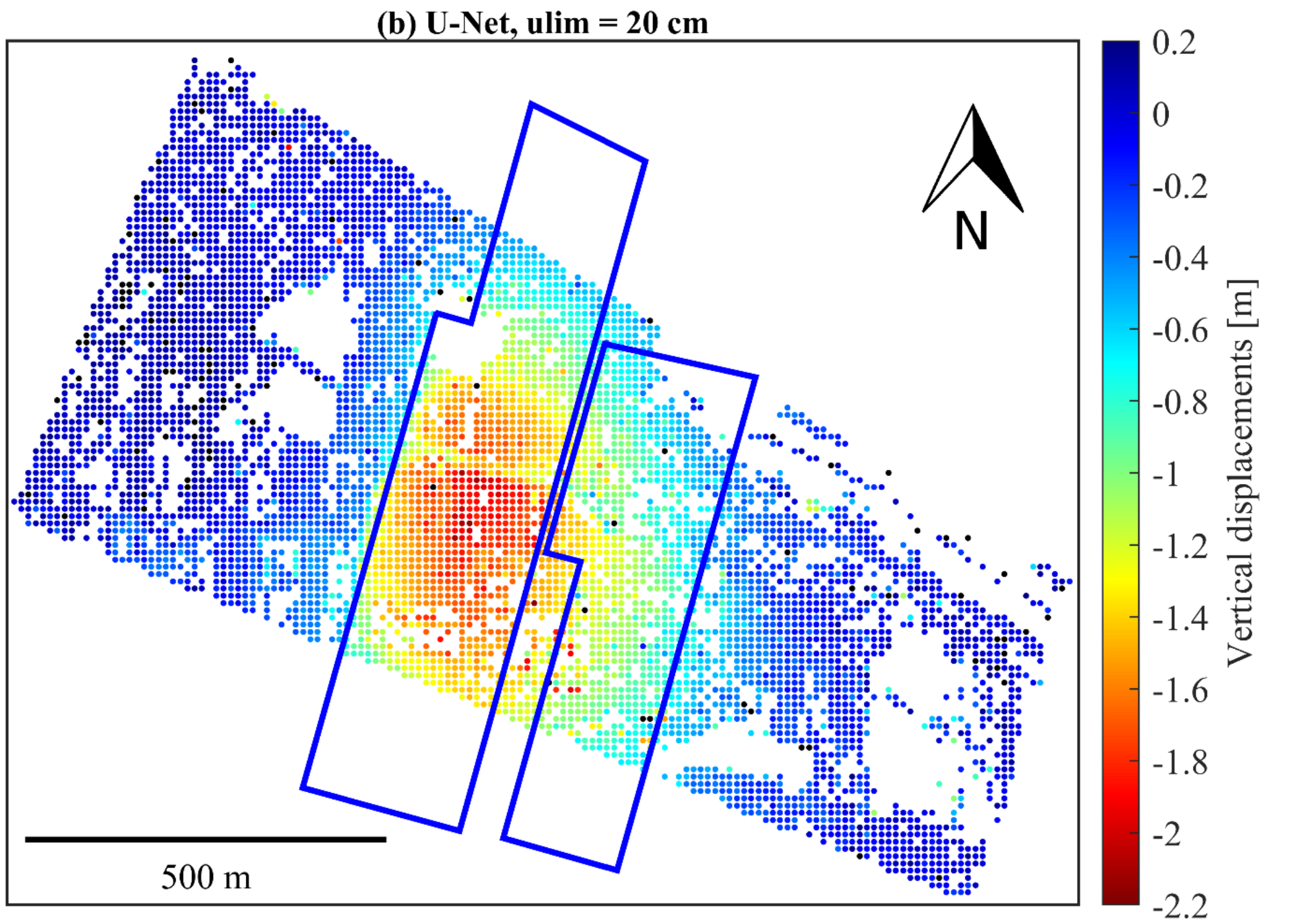

**Figure 9.** Comparison of results for variants w/o filtering and using the U-Net-based algorithm for Jaworzno 2019.07.–2021.11. interval. The underground mining longwall contours are marked in blue.

## 4. Discussion

The developed U-Net model does not offer the same universal applicability as conventional ground filtering algorithms. However, for the tested datasets, the U-Net-based algorithm proved to be more effective in determining vertical displacements than the compared ground filtering algorithms in most cases. This is likely due to the specific characteristics of the study sites, which feature low terrain slopes, limited surface complexity, and relatively dense vegetation coverage. However, these datasets originate directly from or share characteristics similar to those of agricultural areas in the USCB.

Taking into account the limitations connected with machine learning models, extending the application of this algorithm to more complex terrain would require at least additional testing and likely expanding the training and validation datasets. The U-Net-based algorithm has a small number of parameters that need to be set, and their interpretation is straightforward. The main required parameter is ulim, which defines the maximum allowable uncertainty in elevation corrections. The low error dispersion (Figures 6 and 8) and low RMSE of vertical displacements are also important (Tables 2 and 3). Unlike ground filtering algorithms, which perform classification, the U-Net model estimates elevation corrections dynamically. This allows for continuous adjustments rather than strict classification, improving accuracy. The improvement is based on statistical relationships between neighboring points/fragments in the point cloud.

The vertical displacements determined using the U-Net-based algorithm contain a small percentage of outliers, but many of these exhibit a salt-and-pepper noise pattern similar to image artifacts. Such outliers are relatively easy to filter, for example, using moving window techniques combined with Median Absolute Deviation [31], Interquartile Range (IQR) [32], or RANSAC-based methods [33]. Only after applying such filtering techniques can these displacement maps be used for interpolation, subsidence modeling, and analysis of the development of the subsidence basin caused by underground mining operations.

Automatic outlier filtration methods are valuable tools for data cleaning; however, the final selection of a specific method (or combination of methods) and its parameters should be determined experimentally for each case. For example, at the Jaworzno site, during the period 2019.07.–2021.11., a filter based on a moving window and an IQR rule, with a filtering window size of 150 m x 150 m and a filtering window step size of 10 m, yielded satisfactory results. Its application enabled the removal of a significant portion of outlier observations without erroneously removing genuine displacements in areas with non-standard displacement patterns, such as the anomalies on the western slope and in the northern part of the subsidence basin (Figure 10). This filtering had a negligible impact on the quality indicators of the subsidence determination method used in this article, i.e., an RMSE = 10.3 cm, 83.2% nodes with determined displacements, and 0.4% gross errors; any visual changes in the resulting image are subtle and require detailed comparison with Figure 9. Nevertheless, such filtering could be of fundamental importance for subsequent interpolation and the determination of deformation model parameters.

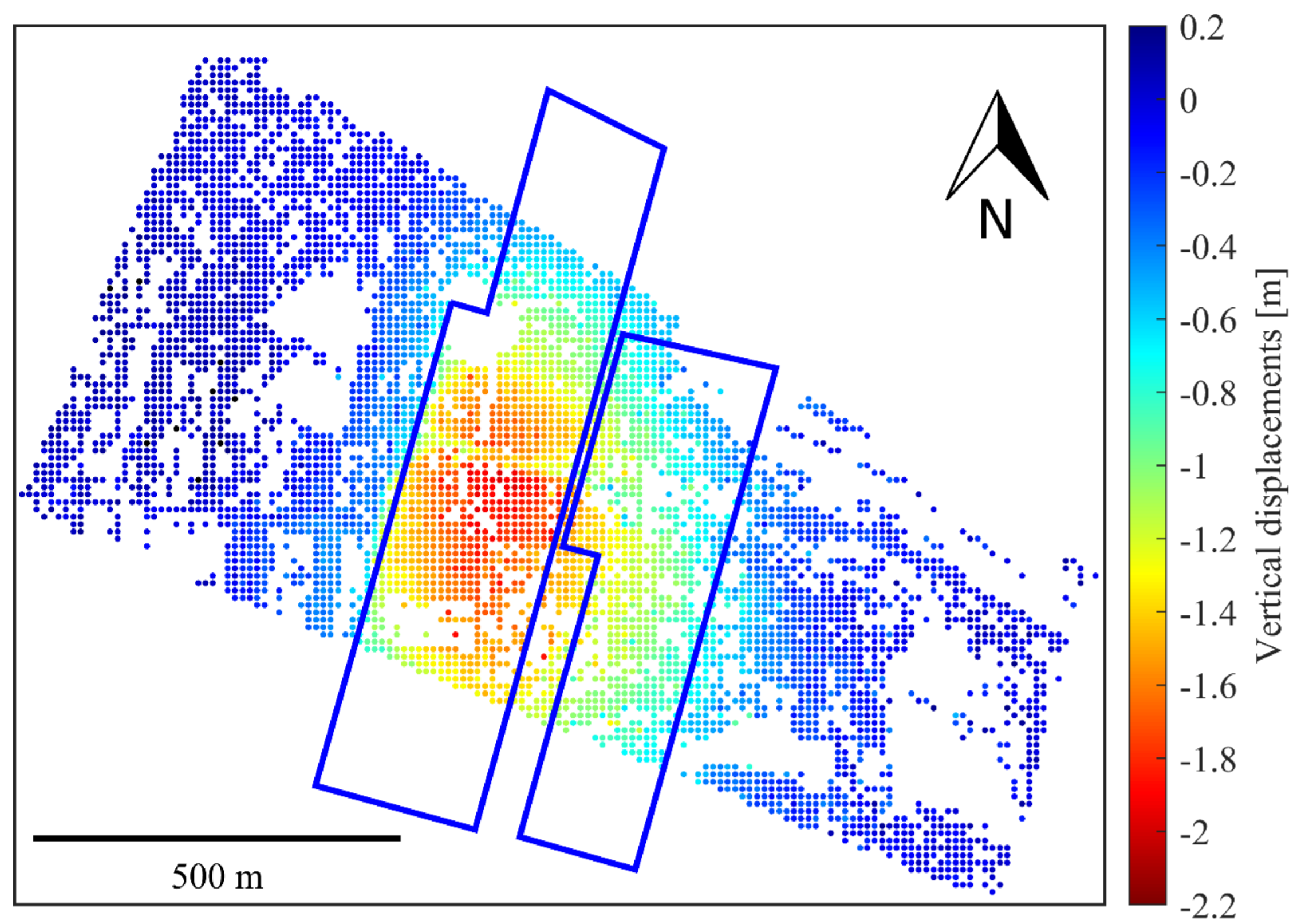

**Figure 10.** Vertical displacements determined using U-Net-based algorithm for Jaworzno 2019.07.–2021.11. interval after IQR-based outlier removal. The underground mining longwall contours are marked in blue.

Among the ground-filtering-based algorithms, the ATIN-based algorithm generally produced the best results, while the SMRF-based algorithm performed slightly worse in most cases. Both of these algorithms (especially ATIN) and the U-Net-based algorithm demonstrated sufficient accuracy for assessing the risk of vertical surface displacements in agricultural areas caused by underground mining operations [34]. However, the accuracy of these methods is insufficient for directly determining the extent of mining influence on agricultural land. Such an assessment would require additional modeling or supplementary observations obtained through alternative methods, such as PSInSAR.

The CSF-based algorithm did not produce satisfactory results in tested scenarios, which was somewhat unexpected, given its popularity and established position in the field. These poor results are likely related to the intensity of vegetation in the analyzed region rather than the terrain's morphology. It is also important to note that ground filtering algorithms are typically designed for ALS-derived point clouds. A review of point clouds filtered using the CSF for the analyzed sites revealed that the algorithm effectively classified buildings and trees (located outside analyzed regions). However, for vegetation in agricultural fields, the algorithm was almost entirely ineffective.

It is important to emphasize that the results presented are representative of the specific conditions under which they were obtained—namely low-slope terrain and agricultural areas. These findings should not be directly extrapolated to different environments, such as mountainous regions or areas with a low vegetation density. However, the ATIN algorithm has also been proven effective in other conditions [3], which serves as a key indicator of its robustness and effectiveness.

For three of the study sites, subsidence accuracy assessment was conducted under the assumption of no displacement, while for the fourth site, accuracy was evaluated using interpolation based on reference measurements at unmarked points. When assessing subsidence estimation accuracy in agricultural areas, these two approaches appear to be significantly better than the widely used method based on comparisons with reference measurements at marked points. Reference measurements at marked points allow for accuracy estimation at characteristic locations, which is appropriate for low-vegetation areas. However, due to potential crop damage, even measurements at unmarked points can typically be conducted only in meadows or fallow lands but not in grain fields or other intensive agricultural crops. This severely limits the applicability and representativeness of such an approach. From this perspective, the most reliable method for assessing accuracy in intensively cultivated areas is to use regions where the surface does not undergo displacement. This approach to accuracy assessment is only valid for areas with similar characteristics to the terrain where actual subsidence occurs. Nevertheless, especially when the estimated subsidence errors are small, this method provides the most reliable accuracy assessment for areas covered by intensive agricultural crops.

## 5. Conclusions

The developed algorithm, based on heteroscedastic regression using a U-Net network, demonstrated high effectiveness in determining vertical ground surface displacements in agricultural areas of the USCB and regions with similar characteristics. In most cases, it ranked at the top of the TOPSIS-based evaluation. The achieved accuracy of estimated vertical displacements is sufficient for assessing the risk of subsidence in agricultural areas affected by underground mining operations. However, it is not precise enough to comprehensively assess the full extent of mining-induced vertical ground surface displacements.

Among the ground filtering algorithms tested, the ATIN-based algorithm produced the best results, followed by the SMRF-based algorithm, which performed slightly worse. The CSF-based subsidence determination algorithm proved ineffective for the tested datasets. While correctly classifying buildings and trees, it misclassified medium and low vegetation in fields and meadows.

**Author Contributions:** Conceptualization, W.G.; methodology, W.G.; software, W.G. and E.P.; validation, W.G.; formal analysis, W.G.; investigation, W.G.; resources, W.G., E.P., P.Ć. and W.M.; data curation, W.G., E.P. and P.Ć.; writing—original draft preparation, W.G. and E.P.; writing—review and editing, W.G. and E.P.; visualization, W.G.; supervision, W.G.; project administration, W.G. and P.Ć.; funding acquisition, W.G., E.P., P.Ć. and W.M. All authors have read and agreed to the published version of the manuscript.

**Funding:** This research project was supported by the program “Excellence Initiative—Research University” for the AGH University of Krakow and statutory research of AGH University of Krakow, Faculty of Geo-Data Science, Geodesy, and Environmental Engineering [grant number 16.16.150.545].

**Data Availability Statement:** The datasets presented in this article are not readily available because the data are part of ongoing studies. Requests to access the datasets should be directed to Wojciech Gruszczyński.

**Acknowledgments:** During the preparation of this work, the authors used Grammarly 1.2.194 and ChatGPT 4 in order to refine the language. After using this tool/service, the authors reviewed and edited the content as needed and take full responsibility for the content of the publication.

**Conflicts of Interest:** The authors declare no conflicts of interest.